\documentclass[conference]{IEEEtran}
\IEEEoverridecommandlockouts
\usepackage{cite}
\usepackage{amsmath,amssymb,amsfonts}
\usepackage{algorithmic}
\usepackage{graphicx}
\usepackage{textcomp}
\usepackage{xcolor}
\usepackage{hyperref}
\usepackage[export]{adjustbox}

\def\BibTeX{{\rm B\kern-.05em{\sc i\kern-.025em b}\kern-.08em
    T\kern-.1667em\lower.7ex\hbox{E}\kern-.125emX}}
\begin{document}

\title{Adaptation of the super resolution SOTA for Art Restoration in camera capture images
}

\author{\IEEEauthorblockN{ Sandeep Nagar}  
\IEEEauthorblockA{\textit{Machine Learning Lab} \\
\textit{IIIT Hyderabad}\\
India \\
sandeep.nagar@research.iiit.ac.in}
\and

\IEEEauthorblockN{ Abhinaba Bala }
\IEEEauthorblockA{\textit{CVIT} \\
\textit{IIIT Hyderabad}\\
India \\
abhinababala04@gmail.com}

\and
\IEEEauthorblockN{ Sai Amrit Patnaik}
\IEEEauthorblockA{\textit{CVIT} \\
\textit{IIIT Hyderabad}\\
India \\
patnaiksai1234@gmail.com}
}

\maketitle

\begin{abstract}
Preserving cultural heritage is of paramount importance. In the domain of art restoration, developing a computer vision model capable of effectively restoring deteriorated images of art pieces was difficult but now we have good computer vision state-of-art. Traditional restoration methods are often time-consuming and require extensive expertise.The aim of this work is to design an automated solution based on computer vision models that can enhance and reconstruct degraded artworks, improving their visual quality while preserving their original characteristics and artifacts. The model should handle a diverse range of deterioration types, including but not limited to noise, blur, scratches, fading, and other common forms of degradation. We adapt the current state-of-art for the image super resolution based on Diffusion Model (DM) and fine-tune it for Image art restoration. Our result show that the instead of fine-tunning multiple different model for different kind of degradation, fine-tunning one super-resolution. We train it on the multiple datasets to make it robust. \textcolor{red}{\href{https://github.com/Naagar/art_restoration_DM}{code link}}.
\end{abstract}


\section{Introduction}
Preserving cultural heritage is of paramount importance. While history has preserved countless masterpieces, the ravages of time have left many artworks faded, damaged, or on the brink of disappearance. Traditional restoration methods are often time-consuming and require extensive expertise. By resurrecting damaged or obscured artworks, we breathe life back into these forgotten stories, reviving the narratives that have shaped our collective consciousness. The domain of image art restoration (IR) holds significant importance within the low-level vision discipline, aiming to enhance the perceptual quality of images that have suffered from a wide array of degradation. This intricate task operates as a versatile and interpretable solution to a range of inverse problems, utilizing readily available denoising techniques as implicit image priors, as demonstrated by \cite{zhu2023denoising}. Within the realm of low-level vision research, IR has persistently remained a focal point, contributing substantially to the enhancement of image aesthetics, as evidenced by the work of \cite{li2023diffusion}. 

In the context of deep learning advancements, a plethora of IR methodologies have harnessed the power of datasets tailored for diverse IR challenges, such as super-resolution (DIV2K, Set5, Set14), rain removal (Rain800, Rain200, Raindrop, DID-MDN), and motion deblurring (REDS, Gopro) \cite{li2023diffusion}. Notably, the emergence of diffusion models (DM) has ushered in a new paradigm within generative models, catalyzing remarkable breakthroughs across various visual generation tasks. The diffusion model , as demonstrated by \cite{xia2023diffir}, excels through a sequential application of denoising networks to replicate the image synthesis process. Capitalizing on the exceptional generative prowess of diffusion models, we employ them as a benchmark for image restoration.

Traditional supervised learning approaches hinge upon extensive collections of distorted/clean image pairs, while zero-shot methods predominantly rely on known degradation modes. However, these methodologies encounter limitations in real-world scenarios characterized by diverse and unknown distortions. To address this concern, some researchers have extended diffusion models to accommodate blind/real-world image restoration scenarios by integrating real-world distortion simulations and kernel-based techniques. This expansion seeks to bridge the gap between diffusion models and the complexity of real-world image restoration challenges, offering a potential avenue for more effective applications in practical settings.

\paragraph{About challenge:}
This work is part of the \emph{Competitions @ ICETCI 2023} \href{https://ietcint.com/user/competitions}{link}.

Motivation: By resurrecting damaged or obscured artworks, we breathe life back into these forgotten stories, reviving the narratives that have shaped our collective consciousness. The participants are required to develop an innovative model that can automate the restoration process and ensure the longevity of art pieces for future generations. 

Objective: The objective of the challenge is to design and implement an advanced computer vision model tailored for the restoration of deteriorated art images. Participants are encouraged to explore various techniques, architectures, and training methodologies to
develop a robust and efficient solution.

\section{Related Work}

\paragraph{Traditional image restoration methods:}  Diffusion-based image restoration techniques rely on partial differential and variational methods, grounded in geometric image models. These methods employ edge information in the damaged area to guide diffusion direction, propagating known information to the target region. While effective for minor image damage, they may yield fuzzy results when handling extensive damage or complex textures \cite{qiang2019survey}.

\paragraph{Deep learning based methods:}
Convolutional neural networks (CNNs) possess remarkable capabilities for learning and representing image features, enabling effective prediction of missing image content. The image restoration process primarily relies on supervised learning methods \cite{zhang2017learning}. 

In contrast to CNNs, which face challenges in supervised image restoration learning, autoencoders (AE) are artificial neural networks proficient at unsupervised learning, effectively learning and expressing input data \cite{mao2016image}. AEs tries to regenerate the images from the latent vector and fails to remove the unordered noise/distortion.

The GAN-based based image restoration method is different from the convolution autoencoder based method \cite{goodfellow2014generative}. The GAN-based image restoration method generates the image to be repaired directly through the generator, and the input can be a random noise vector, and the former is through the whole damaged image to generate the repair area. While GANs excel in generating high-quality images, training them can be challenging, primarily due to the complexity of the loss function employed in the training process.

Another branch of deep learning and generative models, normalizing flows (NF) \cite{rezende2015variational} are also used for the image restoration. NFs are based on the invertible CNN layers \cite{nagar2021cinc, kallappa2023finc} but NFs are slow and cost more computation for high quality input images as compared to CNNs, GANs, and VAEs. NFs works better for the debluring due to their inevitability and tractable nature \cite{wei2022deep}.

Image restoration (IR) represents an essential and demanding endeavor within the realm of low-level vision. Its objective is to enhance the perceptual quality of images afflicted by diverse degradation types. Notably, the diffusion model has made remarkable strides in the visual generation of AIGC, prompting a natural inquiry: "Can the diffusion model enhance image restoration?" \cite{li2023diffusion}. Motivated by tis question, we used the Diffusion Model (DM) base super-resolution to solve the problem of art restoration in the images.

\section{Method Description}
In the realm of artistic representation, the integrity of artworks can be compromised by a multitude of factors such as motion-induced disruptions, various forms of noise, application of filters, and even the intrusion of water. This degradation or distortion also extends to the process of encapsulating art within images, further entailing inherent discrepancies within the representation. Consequently, the task of restoring genuine artistic essence and preserving the authentic artifacts presents a formidable challenge due to the reliance on these images as the solitary source of information regarding the artwork.

\begin{figure*}[!h]
    \centering
    \includegraphics[width=0.9\linewidth]{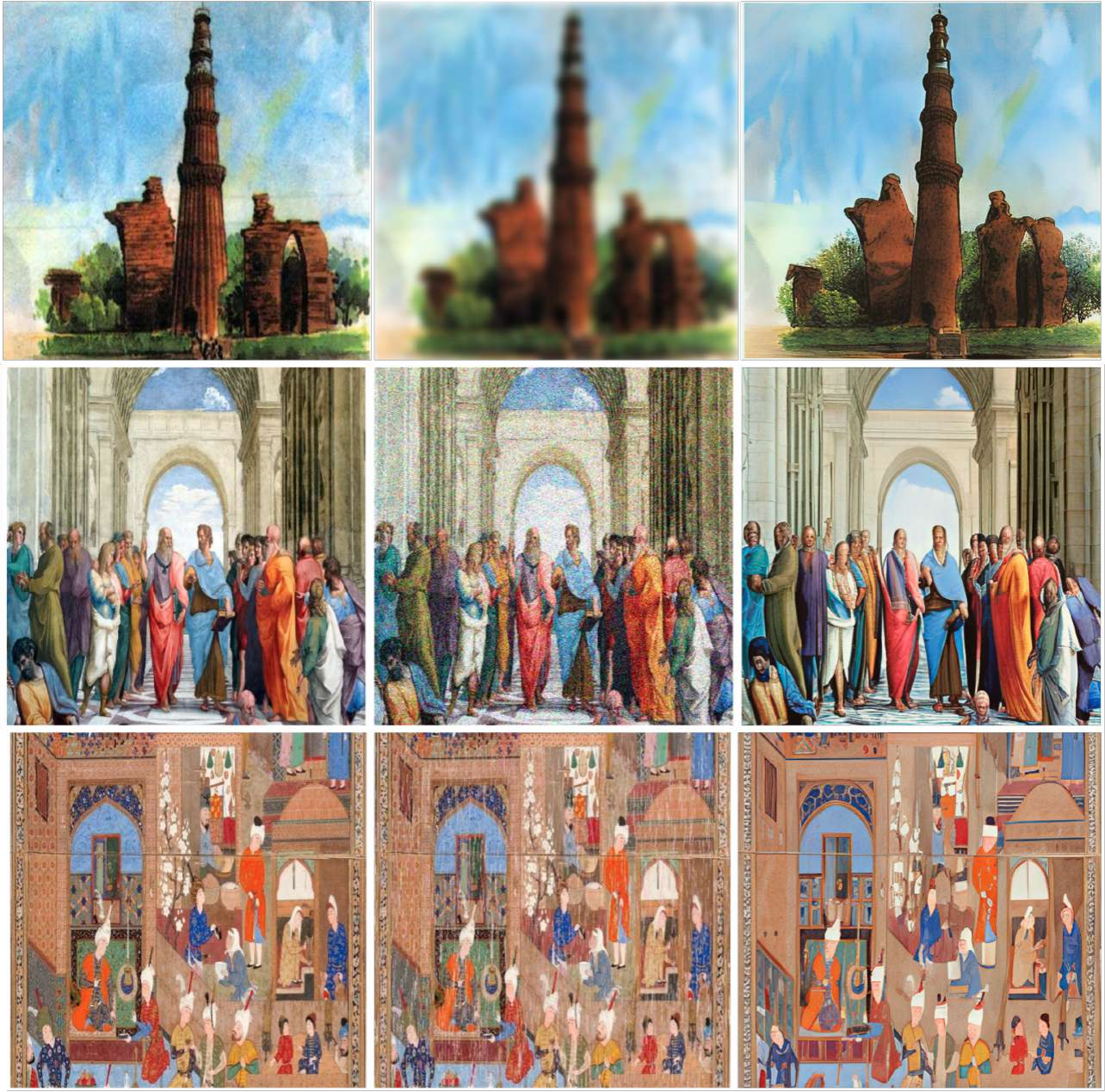}
    \caption{Method: StableSR. Left: GT, Center: distorted by \textbf{noise/ damaged}, Right: Restored}
    \label{fig:sample_fig_label}
\end{figure*}

\begin{figure*}[h]
    \centering
    \includegraphics[width=0.9\linewidth]{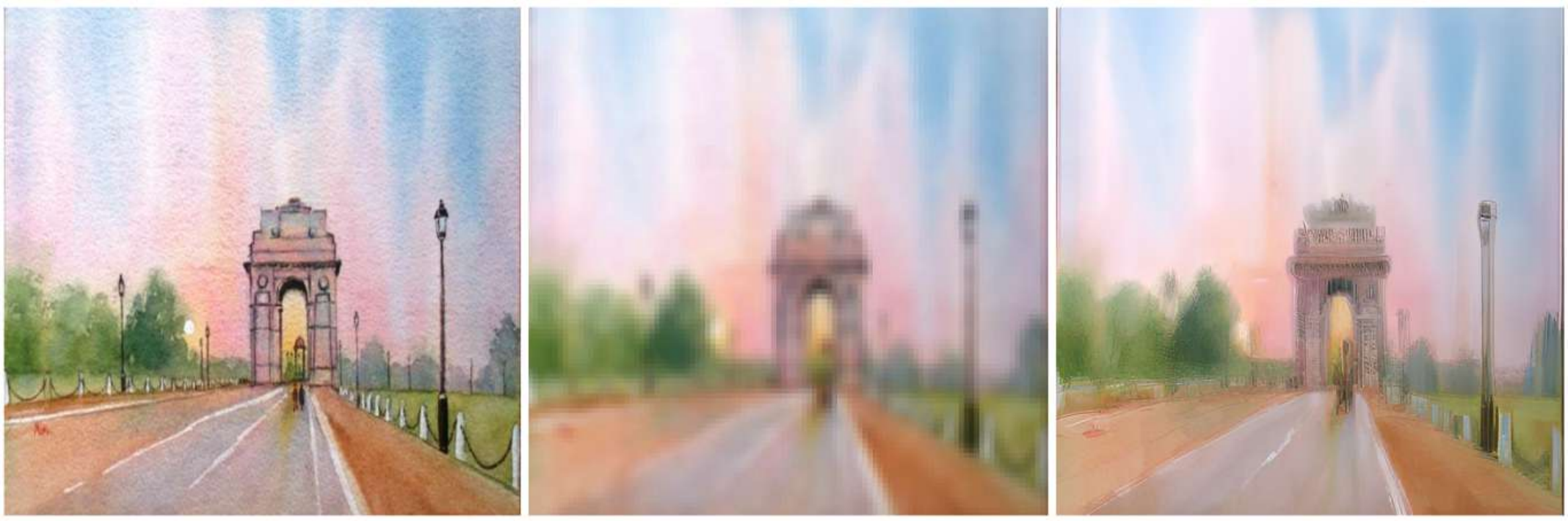}
    \caption{Method: StableSR. Left: GT, Center: distorted by \textbf{pixelating}, Right: Restored}
    \label{fig:sample_fig_label4}
\end{figure*}


The endeavor of rejuvenating impaired artifacts is intricate, marked by its irrevocable nature. This compels the exploration of computer vision models as a recourse, harnessing their capacity to leverage embedded attributes within the images. Among the array of techniques, diffusion models emerge as the paramount state-of-the-art method for both image generation and restoration. In particular, the application of image super-resolution models proves to be salient in the restoration process. These models, in their pursuit of enhanced resolution, inherently address a broad spectrum of prevailing distortions and degradation present in the captured images.

It is imperative to acknowledge that the acquisition of images itself is predisposed to quality deterioration, often stemming from the intricacies of the capturing apparatus. This encompasses the introduction of supplementary noise and filters, thereby exacerbating the challenges intrinsic to preserving the fidelity of art images. 

So we propose to use the super resolution SOTA (StableSR) \cite{wang2023exploiting} to restore the art. Further, we fine tune the StableSR model for the art restoration and reconstruction. Further to verify and compare the StableSR model to other existing super-resolution SOTA, We also test the sample images using the ResShift \cite{yue2023resshift} super resolution model (see fig-\ref{fig:ResShift1}).

\section{Experimental Results}

Within this section, we present the outcomes of our experimentation and conduct a comparative analysis between the ground truth images and their restored counterparts: StableSR, (see Fig-\ref{fig:sample_fig_label}, \ref{fig:sample_fig_label4}). The ensuing paragraphs elaborate on the results obtained through this dual-model approach, shedding light on the performance of StableSR in terms of image restoration. 

\section{Conclusion and Future work}

In conclusion, this endeavor has spotlighted the efficacy of contemporary diffusion models in the realm of image restoration (IR), harnessing their robust generative potential to amplify both structural and textural revitalization. The initial phase of this work entailed leveraging pre-trained weights to establish a foundational baseline, followed by the progressive evolution of the diffusion model for IR applications, with a specific focus on the adaptation of StableSR through a systematic fine-tuning process. This research has further delved into the comprehensive categorization of ten distinct distortions, shedding light on their nuances through the lens of training strategies and degradation scenarios.

Through meticulous analysis, we undertook a comparative assessment of existing works, encompassing both super-resolution and IR domains. Each approach was dissected with precision, affording an intricate taxonomy that delineated their respective strengths and weaknesses. The evaluation process involved an overview of prevalent datasets and evaluation metrics within the diffusion model-based IR landscape. This culminated in a comprehensive comparison of two cutting-edge open-source state-of-the-art (SOTA) methodologies, evaluated through a fusion of distortion and perceptual metrics across three quintessential tasks: image super-resolution, deblurring, and inpainting.

Remarkably, our observations highlighted the effectiveness of training diffusion models on specialized datasets tailored to distinct degradation types. This strategy yielded commendable outcomes, particularly in scenarios mirroring the noise or degradation patterns akin to the training data. As we steer toward future prospects, addressing the challenges inherent in diffusion model-based IR entails exploring diverse baseline datasets and refining training strategies. By doing so, the realm of diffusion models can be further optimized for achieving superior outcomes, marking a promising direction for future exploration. 

\section*{Acknowledgment}

\href{https://ietcint.com/user/competitions}{COMPETITIONS @ ICETCI 2023}


\bibliographystyle{IEEEtran}
\bibliography{IEEEexample}

\begin{thebibliography}{10}
\providecommand{\url}[1]{#1}
\csname url@samestyle\endcsname
\providecommand{\newblock}{\relax}
\providecommand{\bibinfo}[2]{#2}
\providecommand{\BIBentrySTDinterwordspacing}{\spaceskip=0pt\relax}
\providecommand{\BIBentryALTinterwordstretchfactor}{4}
\providecommand{\BIBentryALTinterwordspacing}{\spaceskip=\fontdimen2\font plus
\BIBentryALTinterwordstretchfactor\fontdimen3\font minus \fontdimen4\font\relax}
\providecommand{\BIBforeignlanguage}[2]{{%
\expandafter\ifx\csname l@#1\endcsname\relax
\typeout{** WARNING: IEEEtran.bst: No hyphenation pattern has been}%
\typeout{** loaded for the language `#1'. Using the pattern for}%
\typeout{** the default language instead.}%
\else
\language=\csname l@#1\endcsname
\fi
#2}}
\providecommand{\BIBdecl}{\relax}
\BIBdecl

\bibitem{zhu2023denoising}
Y.~Zhu, K.~Zhang, J.~Liang, J.~Cao, B.~Wen, R.~Timofte, and L.~Van~Gool, ``Denoising diffusion models for plug-and-play image restoration,'' in \emph{Proceedings of the IEEE/CVF Conference on Computer Vision and Pattern Recognition}, 2023, pp. 1219--1229.

\bibitem{li2023diffusion}
X.~Li, Y.~Ren, X.~Jin, C.~Lan, X.~Wang, W.~Zeng, X.~Wang, and Z.~Chen, ``Diffusion models for image restoration and enhancement--a comprehensive survey,'' \emph{arXiv preprint arXiv:2308.09388}, 2023.

\bibitem{xia2023diffir}
B.~Xia, Y.~Zhang, S.~Wang, Y.~Wang, X.~Wu, Y.~Tian, W.~Yang, and L.~Van~Gool, ``Diffir: Efficient diffusion model for image restoration,'' \emph{arXiv preprint arXiv:2303.09472}, 2023.

\bibitem{qiang2019survey}
Z.~Qiang, L.~He, X.~Chen, and D.~Xu, ``Survey on deep learning image inpainting methods,'' \emph{Journal of Image and Graphics}, vol.~24, no.~3, pp. 447--463, 2019.

\bibitem{zhang2017learning}
K.~Zhang, W.~Zuo, S.~Gu, and L.~Zhang, ``Learning deep cnn denoiser prior for image restoration,'' in \emph{Proceedings of the IEEE conference on computer vision and pattern recognition}, 2017, pp. 3929--3938.

\bibitem{mao2016image}
X.-J. Mao, C.~Shen, and Y.-B. Yang, ``Image restoration using convolutional auto-encoders with symmetric skip connections,'' \emph{arXiv preprint arXiv:1606.08921}, 2016.

\bibitem{goodfellow2014generative}
I.~Goodfellow, J.~Pouget-Abadie, M.~Mirza, B.~Xu, D.~Warde-Farley, S.~Ozair, A.~Courville, and Y.~Bengio, ``Generative adversarial nets,'' \emph{Advances in neural information processing systems}, vol.~27, 2014.

\bibitem{rezende2015variational}
D.~Rezende and S.~Mohamed, ``Variational inference with normalizing flows,'' in \emph{International conference on machine learning}.\hskip 1em plus 0.5em minus 0.4em\relax PMLR, 2015, pp. 1530--1538.

\bibitem{nagar2021cinc}
S.~Nagar., M.~Dufraisse., and G.~Varma., ``Cinc flow: Characterizable invertible 3x3 convolution,'' in \emph{The 4th Workshop on Tractable Probabilistic Modeling, Uncertainty in Artificial Intelligence (UAI)}, 2021.

\bibitem{kallappa2023finc}
A.~Kallappa., S.~Nagar., and G.~Varma., ``Finc flow: Fast and invertible k × k convolutions for normalizing flows,'' in \emph{Proceedings of the 18th International Joint Conference on Computer Vision, Imaging and Computer Graphics Theory and Applications - Volume 5: VISAPP,}, INSTICC.\hskip 1em plus 0.5em minus 0.4em\relax SciTePress, 2023, pp. 338--348.

\bibitem{wei2022deep}
X.~Wei, H.~van Gorp, L.~Gonzalez-Carabarin, D.~Freedman, Y.~C. Eldar, and R.~J. van Sloun, ``Deep unfolding with normalizing flow priors for inverse problems,'' \emph{IEEE Transactions on Signal Processing}, vol.~70, pp. 2962--2971, 2022.

\bibitem{wang2023exploiting}
J.~Wang, Z.~Yue, S.~Zhou, K.~C. Chan, and C.~C. Loy, ``Exploiting diffusion prior for real-world image super-resolution,'' \emph{arXiv preprint arXiv:2305.07015}, 2023.

\bibitem{yue2023resshift}
Z.~Yue, J.~Wang, and C.~C. Loy, ``Resshift: Efficient diffusion model for image super-resolution by residual shifting,'' \emph{arXiv preprint arXiv:2307.12348}, 2023.

\end{thebibliography}

\appendix

\begin{figure*}[!h]
    \centering
    \includegraphics[width=0.99\linewidth]{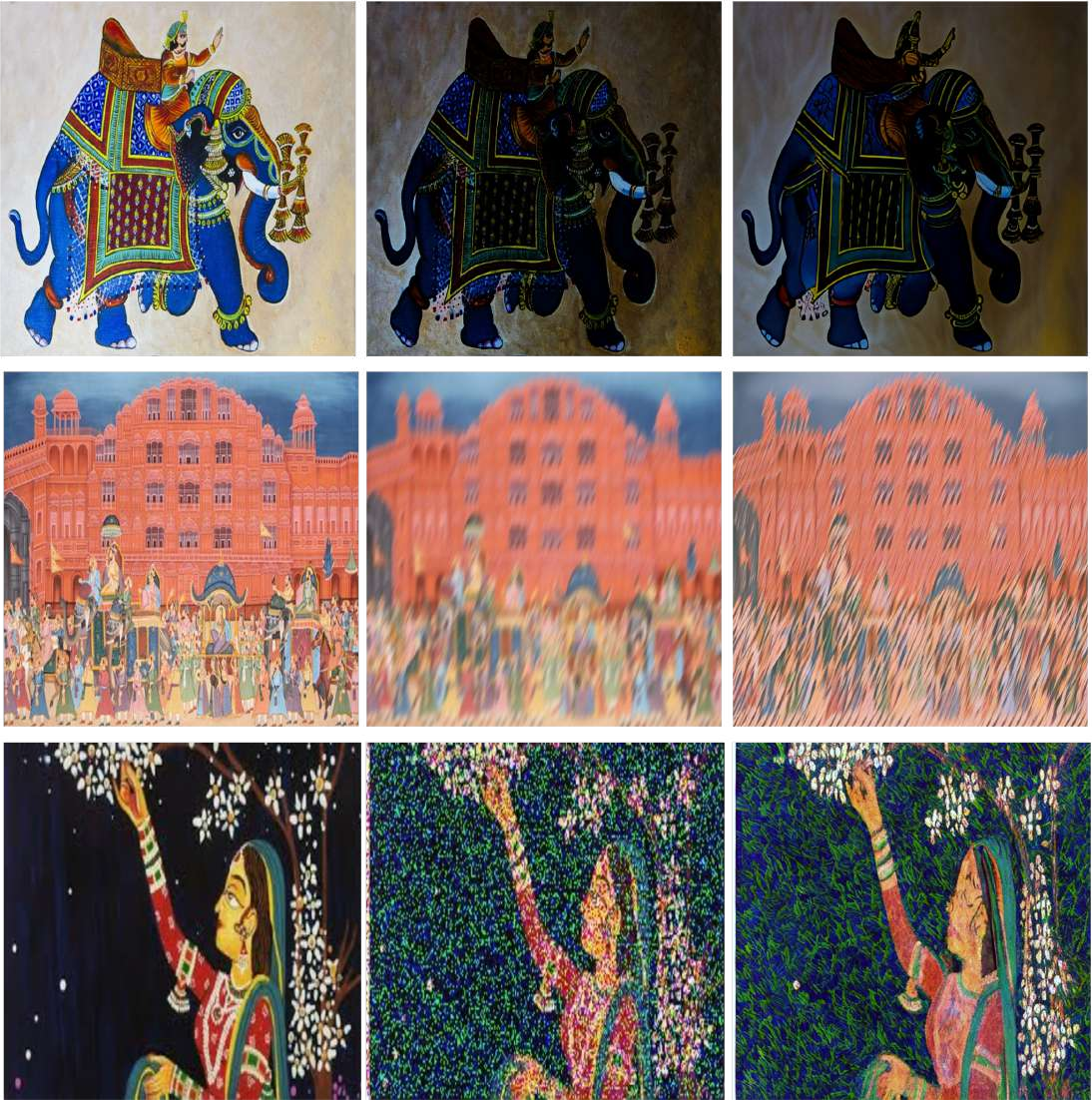}
    \caption{Method: StableSR. Left: GT, Center: distorted by \textbf{damage}, Right: Restored}
    \label{fig:sample_fig_label2}
\end{figure*}

Below, we present additional results where noise becomes integrated into the images, making it difficult to restore the original content (see fig-\ref{fig:sample_fig_label2})

\begin{figure*}[h]
    \centering
    \begin{tabular}{ccc}
        \includegraphics[width=.31\linewidth,valign=m]{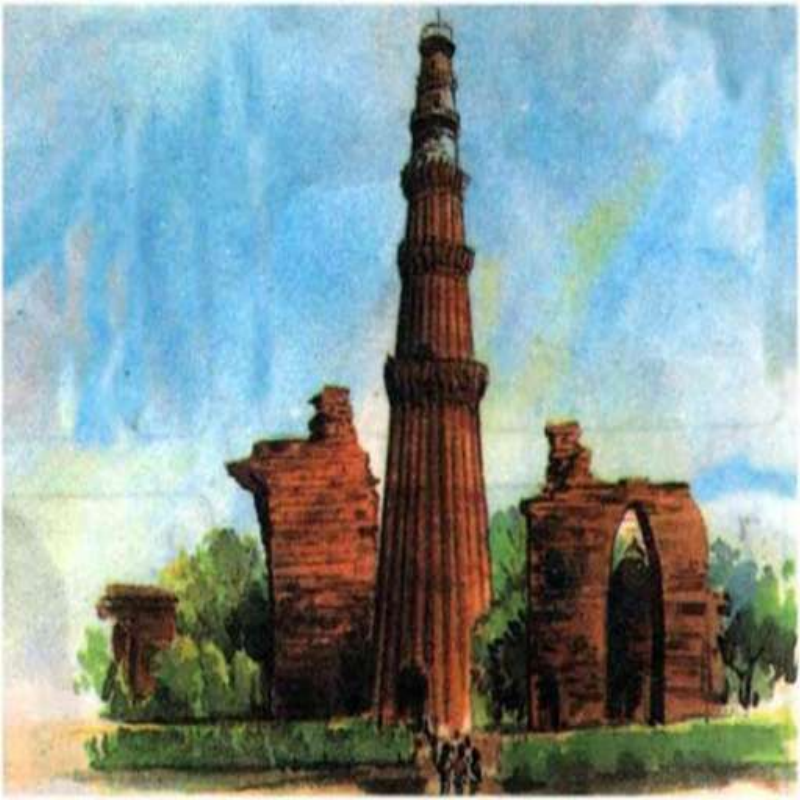} &
        \includegraphics[width=.31\linewidth,valign=m]{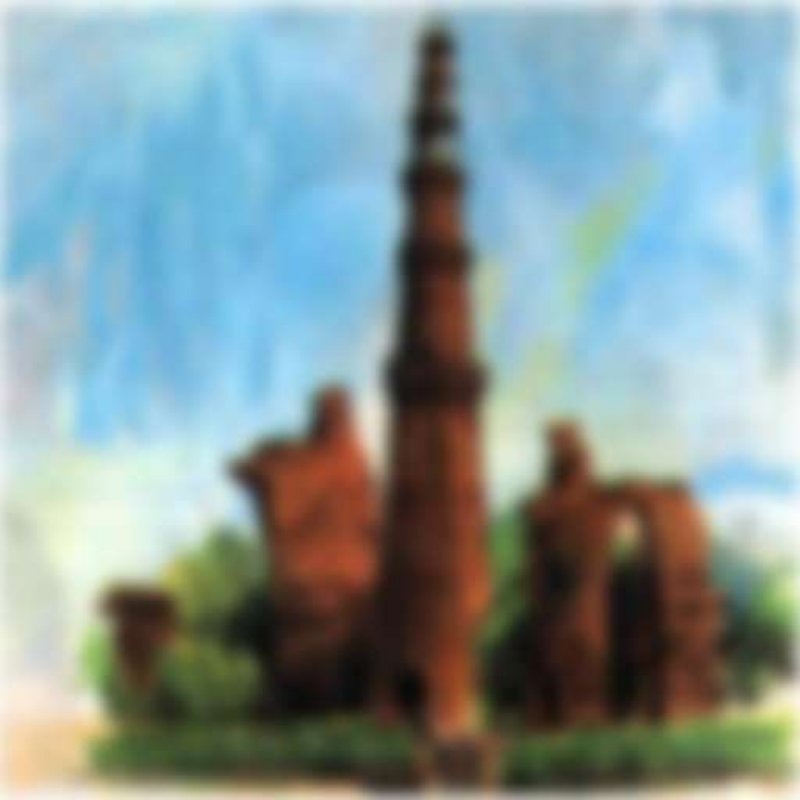} &
        \includegraphics[width=.31\linewidth,valign=m]{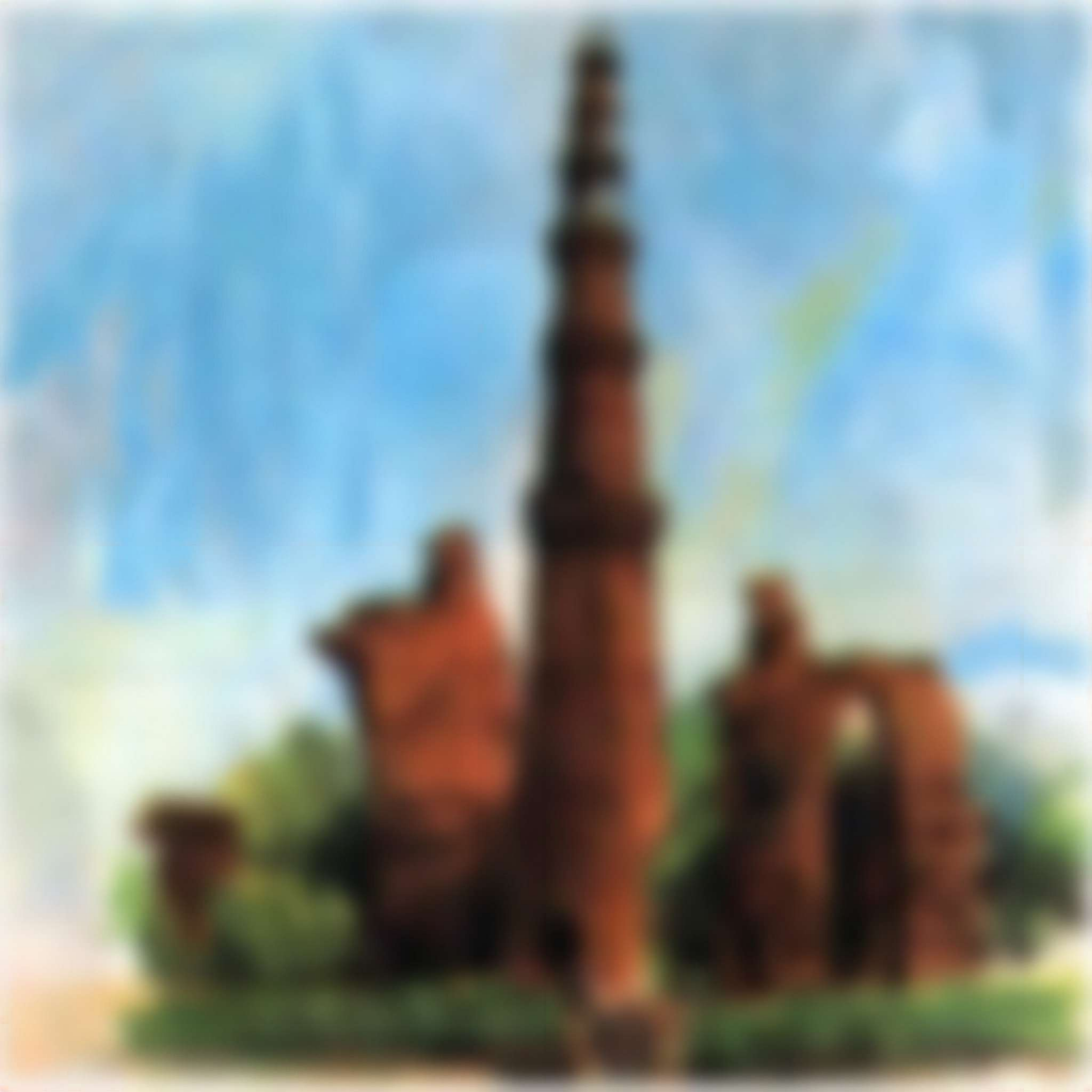} \\
        
        \includegraphics[width=.31\linewidth,valign=m]{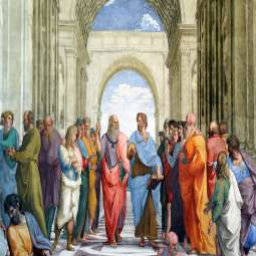} &
        \includegraphics[width=.31\linewidth,valign=m]{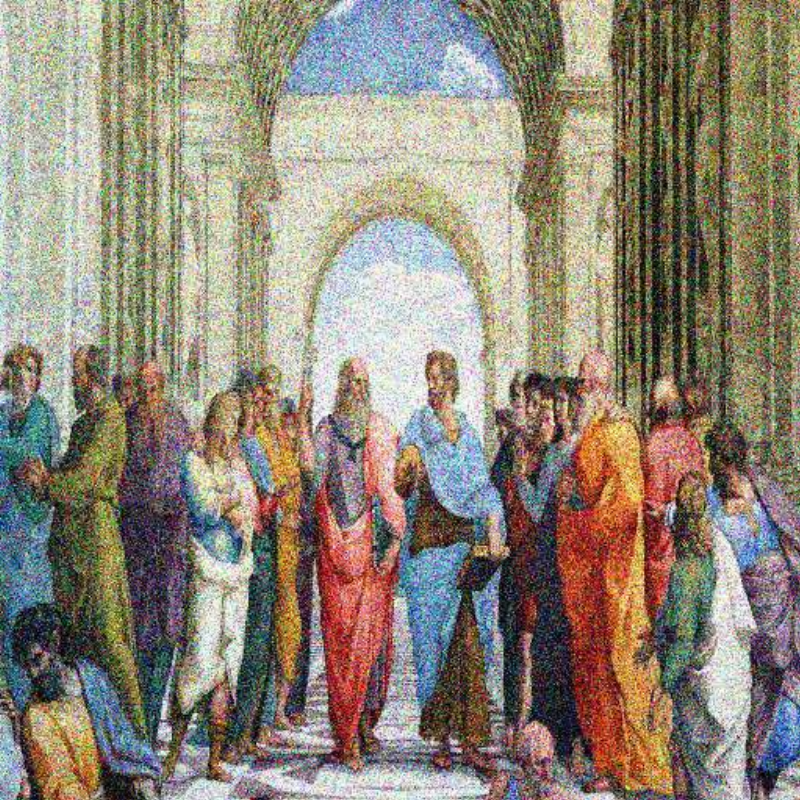} &
        \includegraphics[width=.31\linewidth,valign=m]{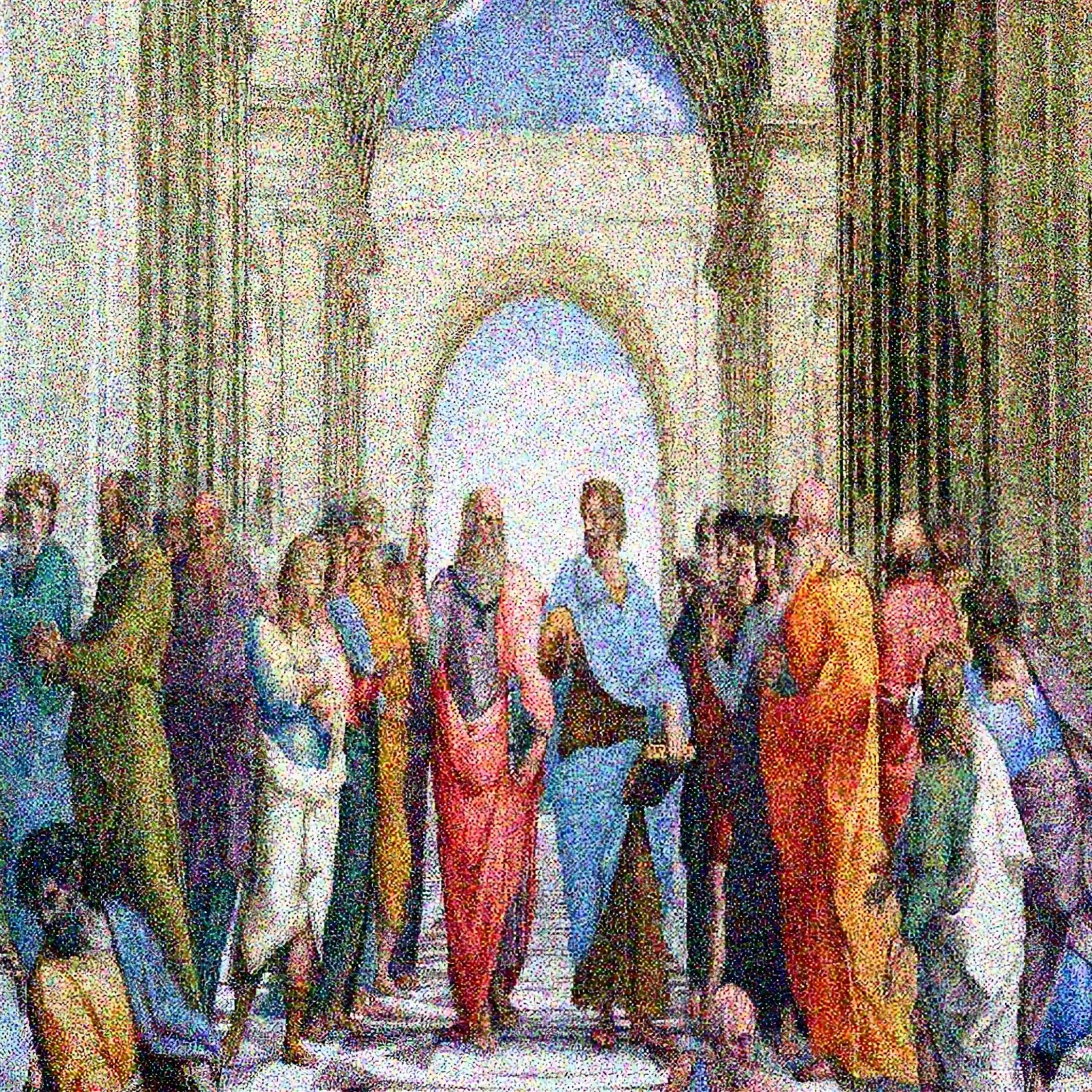} \\
        
        \includegraphics[width=.31\linewidth,valign=m]{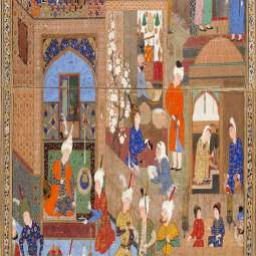} &
        \includegraphics[width=.31\linewidth,valign=m]{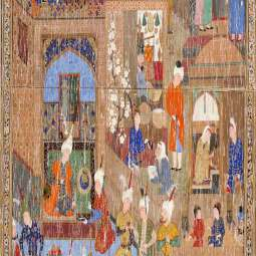} &
        \includegraphics[width=.31\linewidth,valign=m]{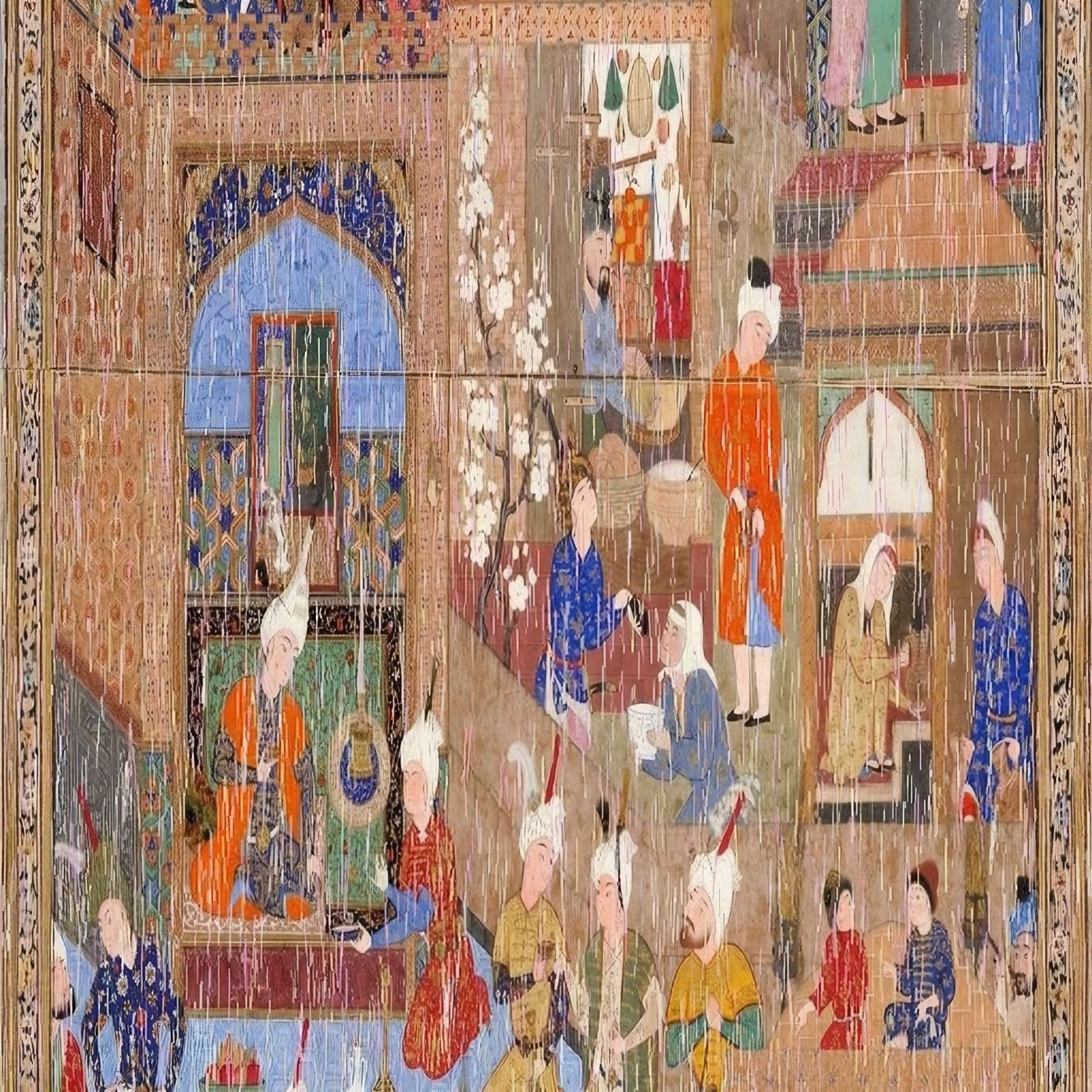} \\
    \end{tabular}
    \caption{Method: ResShift. Left: GT, Center: distorted by artifact, Right: Restored}
    \label{fig:ResShift1}
\end{figure*}

        
        


\end{document}